\documentclass[a4paper, 10pt, conference]{ieeeconf}      
\usepackage{graphicx} 
\usepackage{FG2024}
\usepackage{amsmath}
\usepackage{float}
\usepackage{booktabs}
\usepackage{multirow}
\usepackage{amsfonts}
\FGfinalcopy 

\IEEEoverridecommandlockouts                              
\usepackage{mathptmx} 
\usepackage{amsmath} 
\usepackage{amssymb}  

\def\FGPaperID{7} 

\usepackage{listings}
\usepackage{xcolor}
\usepackage{hyperref}

\definecolor{codegreen}{rgb}{0,0.6,0}
\definecolor{codegray}{rgb}{0.5,0.5,0.5}
\definecolor{codepurple}{rgb}{0.58,0,0.82}
\definecolor{backcolour}{rgb}{0.95,0.95,0.92}

\lstdefinestyle{mystyle}{
    backgroundcolor=\color{backcolour},   
    commentstyle=\color{codegreen},
    keywordstyle=\color{magenta},
    numberstyle=\tiny\color{codegray},
    stringstyle=\color{codepurple},
    basicstyle=\ttfamily\footnotesize,
    breakatwhitespace=false,         
    breaklines=true,                 
    captionpos=b,                    
    keepspaces=true,                 
    numbers=left,                    
    numbersep=5pt,                  
    showspaces=false,                
    showstringspaces=false,
    showtabs=false,                  
    tabsize=2
}

\title{\LARGE \bf
    Aligning Actions and Walking to LLM-Generated Textual Descriptions
}

\author{\parbox{16cm}{\centering
   {\large Radu Chivereanu$^{1}$\thanks{$^{1}$radu.chivereanu@stud.acs.upb.ro},
   Adrian Cosma$^{2}$\thanks{$^{2}$ioan\_adrian.cosma@upb.ro}, 
   Andy Catruna$^{3}$\thanks{$^{3}$andy\_eduard.catruna@upb.ro}, 
   Razvan Rughinis$^{4}$\thanks{$^{4}$razvan.rughinis@upb.ro}, 
   Emilian Radoi$^{5}$\thanks{$^{5}$emilian.radoi@upb.ro},
   }\\
   {\normalsize
   National University of Science and Technology Politehnica Bucharest\\}}%
}

\usepackage{fancyhdr}
\begin{document}

\ifFGfinal
\thispagestyle{empty}
\pagestyle{empty}
\else
\author{Anonymous FG2024 submission\\ Paper ID \FGPaperID \\}
\pagestyle{plain}
\fi
\maketitle

\thispagestyle{fancy}

\begin{abstract}
Large Language Models (LLMs) have demonstrated remarkable capabilities in various domains, including data augmentation and synthetic data generation. This work explores the use of LLMs to generate rich textual descriptions for motion sequences, encompassing both actions and walking patterns. We leverage the expressive power of LLMs to align motion representations with high-level linguistic cues, addressing two distinct tasks: action recognition and retrieval of walking sequences based on appearance attributes. For action recognition, we employ LLMs to generate textual descriptions of actions in the BABEL-60 dataset, facilitating the alignment of motion sequences with linguistic representations. In the domain of gait analysis, we investigate the impact of appearance attributes on walking patterns by generating textual descriptions of motion sequences from the DenseGait dataset using LLMs. These descriptions capture subtle variations in walking styles influenced by factors such as clothing choices and footwear. Our approach demonstrates the potential of LLMs in augmenting structured motion attributes and aligning multi-modal representations. The findings contribute to the advancement of comprehensive motion understanding and open up new avenues for leveraging LLMs in multi-modal alignment and data augmentation for motion analysis. We make the code publicly available at \href{https://github.com/Radu1999/WalkAndText}{https://github.com/Radu1999/WalkAndText}
\end{abstract}

\section{INTRODUCTION}
\label{sec:introduction}
Recently, Large Language Models (LLMs) \cite{zhao2023survey} have gained increased popularity and have proven their applicability in a wide variety of domains beyond natural language processing \cite{wang2023visionllm}. The LLM's expressive power to manipulate language has been proven useful as a method for data augmentation in areas such as image generation \cite{qin2024diffusiongpt}, natural language classification \cite{sahu-etal-2022-data}, search \cite{LLM4IRSurvey}, healthcare \cite{Yuan2024-al,bucur2023utilizing} and others. Furthermore, an emerging approach is leveraging LLMs to generate synthetic data \cite{veselovsky2023generating,li-etal-2023-synthetic}. Synthetic data and automatic annotation processes represent a promising solution for eliminating expensive and time-consuming manual annotations, especially for large scale datasets \cite{cosma22gaitformer}.

Comprehensive motion understanding is a long-lasting problem in the field of artificial intelligence, with definite applications in surveillance, virtual reality, character animation and robotics. Beyond motion generation \cite{actiongpt,motionclip}, previous works have tackled the problem of action recognition in various environments and settings \cite{babel,DBLP:journals/corr/KayCSZHVVGBNSZ17}. An emerging area of research is the problem of zero-shot action classification \cite{ESTEVAM2021159}, in which a model must identify actions that are not contained in the training set, usually using a measure of similarity between motion sequences and textual representations \cite{clip}. Currently, the main difficulty in this task is lack of aligned text and motion data. For example, BABEL-60 \cite{babel}, a popular dataset for motion understanding, offers only single-word labels. More recent resources such as STCM \cite{petrovich2024multi} offer more fine-grained action details in a timeline. 

\begin{figure}[hbt!]
    \centering
    \includegraphics[width=\linewidth]{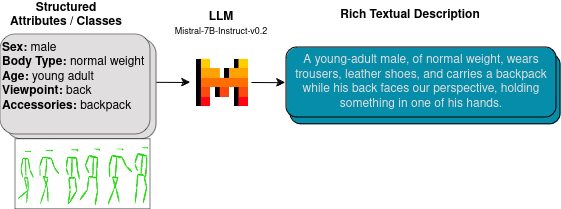}
    \caption{Using the expressivity of Large Language Models, we generate rich textual descriptions for motion sequences across actions and walking sequences. Descriptions are further used to align embeddings between text and motion.}
    \label{fig:text-generation}
\end{figure}

However, only recently methods were developed that make use of the expressivity of LLMs \cite{actiongpt} to augment the multi-modal alignment process. 

In parallel to developments in the area of action understanding, gait analysis \cite{cosma22gaitformer,catruna2023gaitpt} from motion (skeleton) sequences has gained significant traction in biometric identification for uncooperative and in-the-wild settings. The main problem in this domain is disentangling appearance factors from walking patterns to isolate confounding factors and aid person recognition. In this work, we take the opposite direction and attempt to use outward appearance information of a person to retrieve similar walks. Multiple works \cite{cosma22gaitformer,catruna2024paradox} have shown that outward appearance has a definitive impact upon the walking patterns of a person. For instance, Cosma et al, \cite{cosma22gaitformer} incorporated into a gait recognition model person attributes extracted automatically from crops of a person, and obtained positive results. Explicitly connecting textual descriptions and walking sequences represents a new avenue of research, which we are exploring in this work. Intuitively, the way a person is dressed, in terms of clothes and footwear, is subtly affecting walking by both influencing the person's psychology (e.g. walking in a dress or suit compared to walking in a leisurely attire) and by influencing their physical movement (e.g. walking in boots compared to walking in slippers).

In this work, we use LLMs to automatically generate textual description (Figure \ref{fig:text-generation}) of motion sequences, both actions and walking patterns and use them for aligning motion representations to high level linguistic cues. We test this approach on two different tasks and datasets: action recognition using text descriptions on BABEL-60 \cite{babel} and retrieval of walking sequences on DenseGait \cite{cosma22gaitformer} based on appearance attributes. We present qualitative results and show that using LLMs to augment structured motion attributes (e.g. action class or appearance attributes) offers detailed textual descriptions useful in retrieval, as well as quantitative comparisons for action classification and walking retrieval.

This paper makes the following contributions:
\begin{itemize}
    \item We demonstrate the capability of LLMs to synthetically augment structured information about action walking sequences, showcasing high expressivity and enhancing current datasets.
    
    \item We train a CLIP-like model for aligning textual representations with motion sequences in the form of skeletons and obtain comparable results with the state-of-the-art on BABEL-60.

    \item We are the first to explore the feasibility of retrieving skeleton walking sequences from textual descriptions of the outward appearance of the person walking. We showcase a promising avenue for further research on connecting appearance with motion on certain appearance attributes.
\end{itemize}


\section{RELATED WORK}
\label{sec:relatedwork}
Currently, the alignment of human motion to a textual descriptions was mainly explored in works dedicated to motion generation \cite{action2motion,actor,cvae,temos,motionclip,actiongpt}. For instance, Action2Motion \cite{action2motion} and ACTOR \cite{actor} relied on Conditional Variational Autoencoder (CVAE) \cite{cvae} architecture to learn a multi-modal distribution between motion sequences and words describing the action class. Both approaches used a single encoder for both motion and text. TEMOS \cite{temos} shared the same approach, only that it proposed two separate encoders for motion and text which are sharing a latent space conditioned by a reconstruction motion-motion text-motion loss from an universal decoder. Notably, MotionCLIP \cite{motionclip} utilised the latent space of CLIP \cite{clip} in order to derive additional losses to condition the inner representation of the auto-encoder architecture to text. The dataset they used was BABEL, which also has an action recognition benchmark defined. Therefore, in order to prove that they obtain a good alignment between poses and text, they experimented with action classification on their latent space, obtaining good accuracy on BABEL-60 Benchmark \cite{babel}. When incorporating a second loss from the image space of CLIP it further improved results. BABEL-60 dataset was also being utilised in TEACH \cite{teach}, another work for motion generation, but the authors did not explore the downstream task of action classification. 

More similar to us, ActionGPT \cite{actiongpt} uses a LLM to generate coherent sentences of the action labels of each motion sequence, and showed that it improves generalization. They performed a study on enhancing descriptions using LLMs for \cite{teach}, \cite{temos} and \cite{motionclip}. They obtain better results for motion generation with the richer LLM description than the traditional to-the-point information. Zhiyin Shao et al. \cite{unipt} explored using pseudo-text generated from discrete textual attributes of images in a contrastive image-text learning objective. The method proved to be efficient for retrieval.

In the domain of gait, the problem of retrieving gait sequences based on textual descriptions of outward appearance has not been studied in the past. Notably, Cosma and Radoi, \cite{cosma22gaitformer} proposed DenseGait, a large-scale dataset for pretraining in-the-wild gait recognition models, that is also annotated with 42 binary appearance attributes. We use DenseGait to demonstrate a proof-of-concept for gait sequence retrieval from appearance descriptions. Following previous works \cite{actor,actiongpt, cosma22gaitformer}, we generate rich descriptions of actions and person attributes to enable training a CLIP-like model for retrieval of actions and gait sequences from text only.

\section{METHOD}
\label{sec:method}
\begin{figure}[hbt!]
    \centering
    \includegraphics[width=\linewidth]{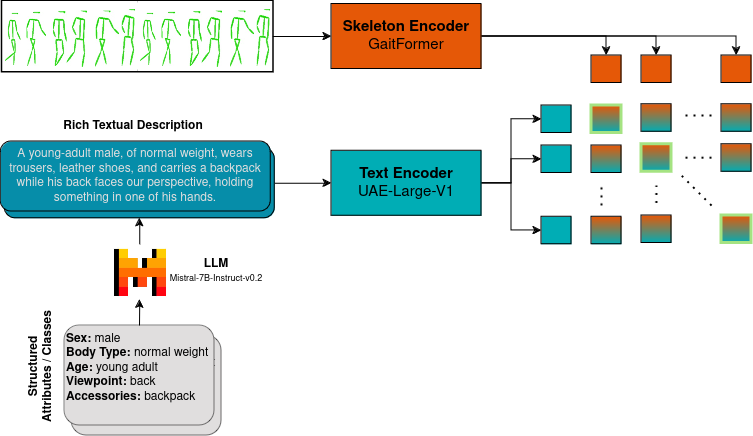}
    \caption{Overall diagram of our method. We use a pretrained large language model to generate rich textual description of a motion sequence. This description is used to align motion representations to their natural language descriptions.}
    \label{fig:gaitclip}
\end{figure}

\subsection{Aligning Motion with Text}
The overall diagram of our method is shown in Figure \ref{fig:gaitclip}. Our architecture receives as input a sequence of poses extracted from a video of a person performing an action and a textual description of the action. The description is automatically augmented through caption generation and given as input to a pretrained text encoder which obtains its embedding. On the other hand, the sequence of poses goes through a different a encoder that extracts the motion-specific features and encodes them into a representation of the same dimensionality as the text embedding. We train the pose encoder to align the representation of the motion to that of the pretrained text encoder. By doing this, the motion encoder learns to project the pose sequences into an embedding space with rich semantic meaning.  

\begin{figure*}[hbt!]
    \centering
    \includegraphics[width=0.75\textwidth]{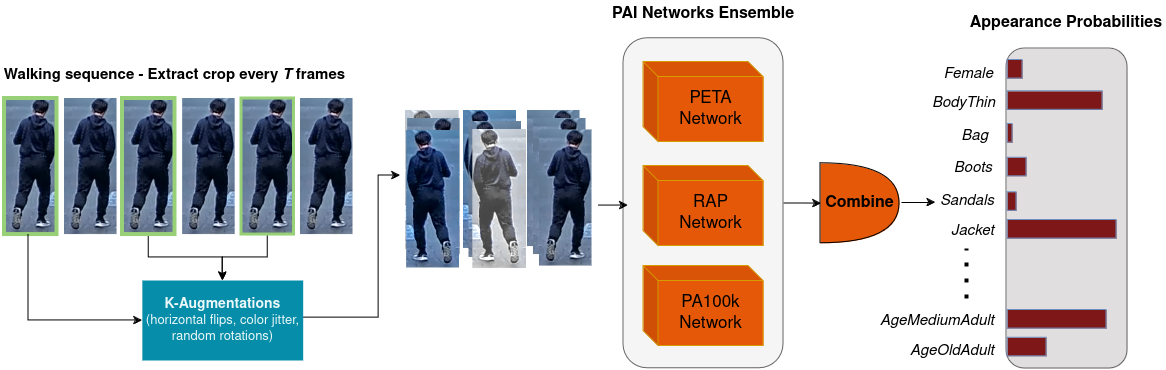}
    \caption{Automatic attribute annotation for walking sequences in the DenseGait. Each walking sequence is augmented 5 times and appearance attributes are estimated using an ensemble of three pretrained models. Figure adapted from Cosma and Radoi \cite{cosma22gaitformer}}
    \label{fig:densegait-annotation}
\end{figure*}

\begin{figure*}[hbt!]
    \centering
    \includegraphics[width=0.75\textwidth]{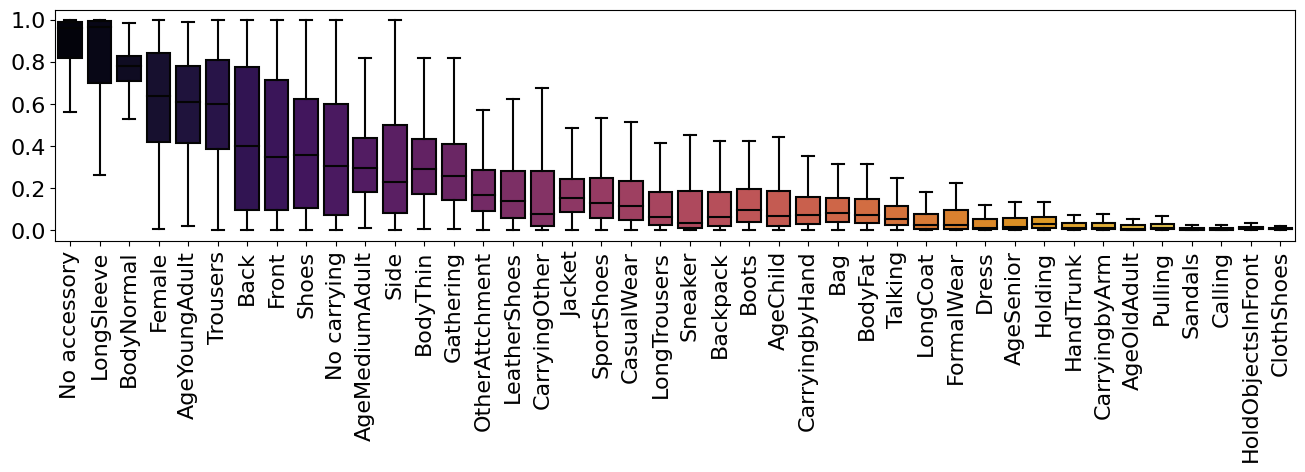}
    \caption{Distribution per appearance feature in DenseGait. Figure adapted from Cosma and Radoi \cite{cosma22gaitformer}}
    \label{fig:densegait-features}
\end{figure*}

Formally, we consider a pose sequence of the form $P = \{p_1, p_2,.., p_n\}$ where $p_k \in \mathbb{R}^{V \times C}$ is the pose obtained from the $k$-th frame in the video and $n$ is the number of poses in the sequence. Our training data consists of pairs of the form $(P_i, T_i)$, where $P_i$ is the pose sequence of the $i$th sample and $T_i$ is its corresponding textual action description. The pose sequence $P_i$ goes through the pose encoder $E_P$ to obtain the latent representation of the motion $z_{P_i}$. At training time, the text label $T$ goes through a pretrained text encoder $E_T$ to obtain the text embedding $z_{T_i}$. 

We train the pose encoder $E_P$ to minimize $d(z_{P_i}, z_{T_i})$ which is the distance in the embedding space between the motion and text representations. The pretrained text encoder remains frozen during training as it already capable of extracting meaningful semantic features from text. In terms of metric for similarity between representations, we experiment with both euclidean and cosine distances.  

\subsection{Datasets}\label{sec:dataset}
Our experiments are conducted on two large-scale datasets for motion analysis: BABEL and DenseGait.

BABEL is a large dataset with language labels describing the actions being performed in MoCap sequences. The dataset is built from a carefully selected and annotated subset of AMASS database.
The full dataset consists of 250 different classes, amounting to a total of  43 hours of mocap sequences. Out of these 250 classes, 60 were selected to define the BABEL-60 benchmark, which our work reports itself to. It contains both sequence and frame level captioning. In our experiments we made use of the sequence labels.

DenseGait is a large-scale gait dataset consisting of automatically obtained walking sequences and their corresponding appearance attributes. The walking sequences were obtained with pose estimation from raw surveillance footage while the appearance attributes were obtained with an ensemble of pretrained models as shown in Figure \ref{fig:densegait-annotation}. The automatically obtained appearance attributes are shown in Figure \ref{fig:densegait-features} and encompass characteristics such as gender, age, clothing, as well as direction. For the gait sequences in DenseGait, we select the set of features with a confidence higher than the threshold value of 0.5. We split the dataset into 3 groups: 80\% for training, 10\% for validation, and the remaining 10\% for testing. Table \ref{tab:densegait-split} shows the number of sequences for each dataset split and its corresponding unique features.

\begin{table}[hbt!]
 \caption{DenseGait subsets utilized in our experiments.}
    \centering
    \begin{tabular}{c|cc}
        \textbf{Split} & \textbf{Number of Sequences} & \textbf{Unique attribute combinations} \\
        \toprule
         Train & 174363 & 7401 \\
         Test & 21795 & 2655 \\
         Val & 21795 & 2721 \\
    \end{tabular}
    \label{tab:densegait-split}
\end{table}

\subsection{Label Augmentation through Caption Generation}
As shown in other works \cite{jia2021scaling,betker2023improving} on aligning text with other modalities, problems can arise from the poor quality of the text descriptions. This is also our case as the experimental dataset \cite{babel} contains text descriptions of the actions that are too concise and superficial. Instead of describing all the movements of the person, the description is either too vague or a few words encompassing the global action. This can limit the expressivity of the trained pose encoder as its training signal is based on text embeddings that do not hold enough information. Ensuring the quality of descriptions demands extensive manual effort leading to significant costs.

To solve this problem we propose improving the text description by employing LLMs to generate more detailed descriptions of the actions. By doing this, the training signal of the pose encoder is more expressive, allowing for a more complex understanding of the movement. The description were generated with a prompt inspired by \cite{actiongpt}: \textit{"Describe in detail a person’s body movements who is performing the action: \{ACT\}"}.

\subsection{Generating Appearance Descriptions for Walking Sequences}

One of the shortcomings of label augmentation is that, intrinsically, the correlation between movement and its generated description is based only on the generic attributes of the depicted action. The quality and diversity of the generated description in is influenced by the number of initial descriptors that are used as guidance for the LLM. Using DenseGait, which provides a set of 42 binary appearance attributes for each skeleton sequence, we can achieve a much more detailed description of the motion sequence. For generating descriptions for the appearance attributes, we used Mistral-7B-Instruct-v0.2 \cite{jiang2023mistral}.

The prompt instructs the model to provide a description for a person having a particular set of appearance attributes: \textit{"Concisely describe a person with the following features: \{features\}"}. The features are injected in json format.  Using "concisely", we limit the risk of the model drifting from the instructions.

\subsection{Model Architectures}

For the skeleton sequence encoder, both for encoding gait and action sequences, we used the GaitFormer architecture \cite{cosma22gaitformer}, which is comprised of a transformer encoder that operates on a sequence flattened skeletons. The implementation is based on the ACTOR encoder, which adopts the continuous 6D representation \cite{rot6d}.

For the text encoder, we used a pretrained UAE-Large-V1 \cite{uae}, a model that achieves state-of-the-art results on Massive Text Embedding Benchmark (MTEB) \cite{mteb}. During training, the text encoder is frozen, and only the skeleton encoder is trainable.

\subsection{Training Objectives}
For training the pose encoder to obtain an embedding space with rich semantic meaning we experiment with multiple types of losses that aim to minimize the distance $d(z_{P_i}, z_{T_i})$.

Our first experimental training objective follows the approach of CLIP \cite{clip} which uses the contrastive loss to minimize the cosine distance between image and text representations. We adapt the contrastive approach to our case of motion and text representations. Given a mini-batch, the pose encoder is trained to maximize the cosine similarity between the positive pairs of pose and text representations while minimizing the cosine similarity between all the negative pairs. However, certain samples can have different textual description but with similar semantic meaning. Because of this, there are scenarios in which positive pairs are incorrectly treated as negative, affecting the overall training. To address this, we compute the similarity between all text descriptions and mask out all negative pairs over a certain threshold.

The second approach involves training the model to only minimize the euclidean distance between the motion and text representations. For this, we employ the Mean Square Error (MSE) which can be computed as:

\begin{equation}
    L_{MSE} = \frac{1}{N} \sum_{i=1}^{N} (Z_{P_i} - Z_{T_i})^2
\end{equation}

The final approach also minimizes the euclidean distance but with the triplet objective which takes into account negative pairs as well. We consider as anchor the embedding of the motion $Z_{P_i}$ and as positive its text representation $Z_{T_i}$. To obtain the negative embedding, we randomly sample with uniform probability a different text embedding from the other classes. For DenseGait, a class is defined by the set of attributes of a given movement sequence. As in the case of the contrastive objective, if the similarity of the positive and sampled negative is over a certain threshold, we omit it in the computation. The triplet objective can be formulated as:
\begin{equation}
    L_{Triplet} = max(d(z_{P_i}, z_{T_i}) - d(z_{P_i}, z_{T_j}) + m, 0)
\end{equation}
where d is the euclidean distance, m is the enforced margin and j is the index of the negative sample.

\subsection{Implementation Details}
We employ the GaitFormer variant with a depth of 12 attention layers and a hidden size of 768. The input for the model is a motion sequence consisting of 60 poses obtained from consecutive frames. For action recognition on BABEL-60 we train the model with AdamW optimizer, a batch size of 80, a learning rate of 1e-05 for 200 epochs. In our experiments in retrieval and classification on DenseGait, we employ a batch size of 256 for 140 training epochs. All our experiments are conducted on an NVIDIA A10 GPU.

\section{EXPERIMENTS}
\label{sec:experiments}
\subsection{Generating descriptions for motion}

\begin{table}[hbt!]
\caption{Example of generated descriptions for BABEL-60 action recognition benchmark.}
\label{tab:action-descriptions}
\begin{center}
    \resizebox{\linewidth}{!}{
    \begin{tabular}{c|p{7cm}}
    \textbf{Label} & \textbf{Generated description (partial)}\\
    \toprule
    \multirow{4}{*}{run} & Running is a complex action that involves the entire body. \\
    & Starting from the head, it is typically held high and \\
    & facing forward, following the direction of movement. \\
    & The eyes focus on the path ahead to anticipate any obstacles. \\
    & The upper body leans slightly forward to propel the body in...\\
    \midrule
     \multirow{4}{*}{turn} & The action "turn" involves a sequence of body movements,\\
    & The person shifts their weight onto the foot in the direction\\
    & The person begins to pivot on the ball of their weighted foot\\
    \midrule
    \multirow{2}{*}{sit} &...they start bending at the knees and hips simultaneously...\\ 
    &...Their thighs, shins, and feet adjust to the shift in weight...\\
     \midrule
     \multirow{2}{*}{lift something} &The person would first approach the object and stand \\ 
    &close to it, with feet shoulder-width apart for balance and stability... The individual would then squat down by bending at the hips and knees... As they squat down, the person would reach for the object... \\
    \midrule
     \multirow{2}{*}{squat} &Performing a squat involves a series of precise body movements... Their chest remains lifted, and the spine maintains a neutral position. The person then lowers their body towards the ground, as if sitting back into an imaginary chair. The knees should track in line with the feet, not caving inwards or bowing out. The person continues to lower their body until their thighs are parallel, or almost parallel, to the ground...To rise, the person pushes down through their heels, straightens their knees and hips simultaneously, and lifts their body back to the starting position...\\ 
    \end{tabular}}
\end{center}
\end{table}

In our experiments, we employ Large Language Models to generate more detailed textual descriptions of the labels with the purpose of improving the contrastive training method.

For BABEL-60, the generative model was instructed to provide a detailed analysis with intricate details for a given action class. We selected a subset of examples in Table \ref{tab:action-descriptions}. It can be observed that the initial label is superficial, consisting mostly of one or two words, whereas the generated descriptions are more rich and explain the sequence of movements.

We further show in Section \ref{subsec:action_recon}, evidence indicating that the descriptions produced offer a richer semantic context for our model.

Since the description generation process relies solely on the action label as a reference point, it results in a small set of generic descriptions for the entire dataset, neglecting a significant portion of information that may be encoded in the skeleton. To address this constraint and enhance the quality of motion-text pairs correlation, we use the DenseGait dataset.

DenseGait provides a selection of textual annotations that correspond to the appearance attributes for each individual movement sequence. This enables the LLM to generate rich textual descriptions with strong correlation to the motion frames. We show the initial attributes and their corresponding generated description in Table \ref{tab:gait-descriptions}.

\begin{table}[hbt!]
\caption{Selected appearance descriptions for DenseGait in-the-wild gait dataset.}
\label{tab:gait-descriptions}
\begin{center}
    \resizebox{\linewidth}{!}{
    \begin{tabular}{p{3cm}|p{7cm}}
    \textbf{Attributes} & \textbf{Generated description (partial)}\\
    \toprule
        sex: female,
        bodytype: normal,
       age: young-adult,
       viewpoint: back,
       actions: 'gathering' & \textit{A young-adult female with a normal body type is standing with her back facing you, engaged in the action of gathering something.}\\
    \midrule
        sex: male,
        bodytype: normal weight,
        age: young-adult,
        viewpoint: back,
        accessories: backpack,
        actions: carry something in hand & \textit{A young-adult male, of normal weight, is seen from the back, carrying a backpack and holding something in one hand.}\\
    \midrule
    sex: male,
   bodytype: normal weight,
   viewpoint: back & \textit{A normal weight male, viewed from the back, is an individual with an average build. He has a balanced proportion of body weight and height, with a toned and healthy appearance. His posture is likely upright, with his shoulders even and his back straight. His body shape is neither too thin nor too heavy, giving him a fit and athletic appearance.}\\
     \midrule
    sex: male,
   bodytype: normal weight,
   age: adult,
   viewpoint: back,
   clothing style: casual,
   footwear: sandals & \textit{A middle-aged man with a normal weight body stands before you, his back facing you. He is dressed in casual attire, and his feet are clad in sandals.} \\
    \midrule
    sex: male,
    bodytype: normal weight,
    age: young-adult,
    viewpoint: front, & \textit{A young-adult male, of normal weight, is standing before you. He wears a pair of fitted trousers and a top that isn't specified in the given data. His feet are clad in sneakers, completing his casual attire.}\\ 
    \end{tabular}}
\end{center}
\end{table}

\subsection{Comparison with Action Recognition}
\label{subsec:action_recon}

We evaluate the features learnt by the pose encoder on the BABEL-60 benchmark in the context of Action Recognition. We obtain the action prediction by computing the distance between the text embeddings of the labels and the embedding obtained from the pose encoder.

\begin{table}[hbt!]
\caption{Results on action recognition on BABEL-60 benchmark. }
\label{tab:sota-action-results}
    \begin{center}
    \resizebox{\linewidth}{!}{
    \begin{tabular}{ll|cc}
        \textbf{Method} & \textbf{Label Text Type} & \textbf{Top 1 (\%)} & \textbf{Top 5 (\%)} \\
        \toprule
        MotionCLIP \cite{motionclip} & Words & 35.50 & 57.71 \\
        2s-AGCN \cite{2sAGCN} & - & 41.14 & 73.18 \\
        \midrule
        \multirow{2}{*}{Contrastive Cosine}  & Words & 45.39 & 68.00\\
        & Generated Description \textbf{(ours)} & 46.41 & 68.05\\
        \multirow{2}{*}{MSE} & Words & 51.58 & 67.61\\
        & Generated Descriptions \textbf{(ours)} & 51.28 & \textbf{70.00} \\
        \multirow{2}{*}{Triplet Loss} & Words & 52.45 & 64.92\\
        & Generated Descriptions \textbf{(ours)} & \textbf{52.52} & 68.83\\
    \end{tabular}}
\end{center}
\end{table}

Table \ref{tab:sota-action-results} shows the performance of our experimental training methods in terms of Top 1 and Top 5 accuracy. The triplet loss obtains the highest Top 1 accuracy of 52.52\%. We observe a positive impact of generated descriptions mostly in the case of Top 5 metric where, in conjunction with MSE loss, the model achieves 70\% accuracy. Our proposed methods outperform other approaches \cite{motionclip, 2sAGCN} in terms of Top 1 accuracy for action recognition on this benchmark.

\begin{table}[hbt!]
    \caption{Selected exmaples of LLM-generated synonyms for initial classes in BABEL-60.}
    \centering
    \small 
    \begin{tabular}{l|l}
        \textbf{Label} & \textbf{Synonym} \\
        \toprule
        walk & stroll \\
        stand & maintain posture \\
        hand movements & manual gestures \\
        turn & rotate \\
        interact with/use object & manipulate item \\
        arm movements & upper limb movements \\
        t pose & neutral pose \\
        step & stride \\
        backwards movement & retrograde motion \\
        raising body part & elevating body part \\
    \end{tabular}
    \label{tab:action-synonyms}
\end{table}

We explore the zero-shot capabilities of our approach by determining whether the model maintains its ability to distinguish between classes when trained on the set of label names and tested with a new set obtained from synonyms of the initial label names. We employ the LLM to automatically obtain the synonyms, such as the sample shown in Table \ref{tab:action-synonyms}.

\begin{table}[hbt!]
\caption{Results on BABEL-60 benchmark training with labels and testing with generated synonyms.}
\label{tab:synonyms}
    \begin{center}
    \resizebox{\linewidth}{!}{
    \begin{tabular}{l|cc}
        \textbf{Method} & \textbf{Top 1 (\%)} & \textbf{Top 5 (\%)} \\
        \toprule
        Triplet Loss Words & 42.53 & 56.43\\
        Triplet Loss Generated Descriptions & \textbf{47.75} & \textbf{64.04} \\
    \end{tabular}}
\end{center}
\end{table}

The results of the zero-shot synonym experiment are shown in Table \ref{tab:synonyms}. We only report the performance of the triplet loss as it obtained the highest accuracy in previous experiments. The model manages to obtain a comparable accuracy even when processing only the synonym embeddings. Furthermore, generated captions bring a considerable improvement in this zero-shot setting, increasing the Top 1 accuracy by more than 5\% and the Top 5 accuracy by approximately 8\%.

\subsection{Retrieving Walking Sequences based on Appearance Description}

\begin{figure}[H]
    \centering
    \includegraphics[width=\linewidth]{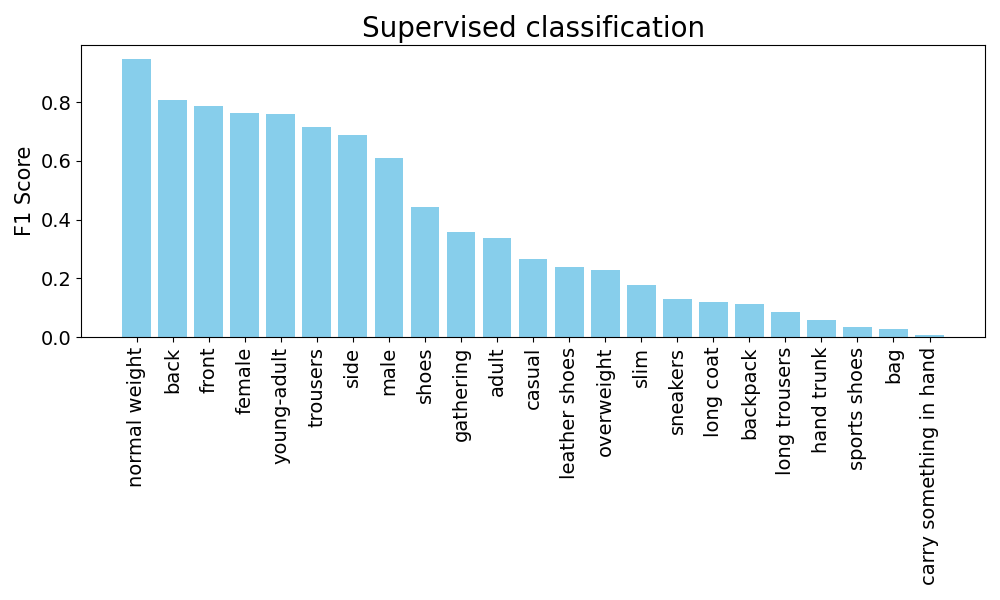}
    \caption{Multi-label classification results for directly predicting the appearance attributes given a walking sequence.}
    \label{fig:classif}
\end{figure}

As a sanity check, to ensure that indeed appearance attributes can be derived from walking sequences, we perform direct multi-label classification of the 42 appearance attributes given a skeleton sequence. We train the initial GaitFormer with binary cross-entropy loss to perform multi-label classification. The results are shown in Figure \ref{fig:classif}. In line with the results from Cosma and Radoi \cite{cosma22gaitformer}, we obtain reasonable results on several attributes. For instance, for walking direction (i.e., back, front, side), gender, age and some clothing and footwear items (i.e., trousers, shoes), the model obtains a reasonable F1 score.

Further, we train a CLIP-like model to align the generated text descriptions with the walking sequences. In this setting, we evaluate the performance for retrieving a gait sequence given a text representation of the appearance. As such, we used the NDCG@K metric. We define relevance (\textit{rel}) for a retrieved element based on the number of matches between the attributes of the retrieved element and the attributes of the retrieved walking sequence.

\begin{equation}
    {DCG@K} = \sum_{i=1}^{K} \frac{rel_{i}}{\log_2(1 + i)}
\end{equation}

\begin{equation}
    {IDCG@K} = \sum_{i=1}^{K} \frac{ideal\_rel_{i}}{\log_2(1 + i)}
\end{equation}

\begin{equation}
    {NDCG@K} = \frac{DGC@K}{IDCG@K}
\end{equation}

\begin{figure}[hbt!]
    \centering
    \includegraphics[width=\linewidth]{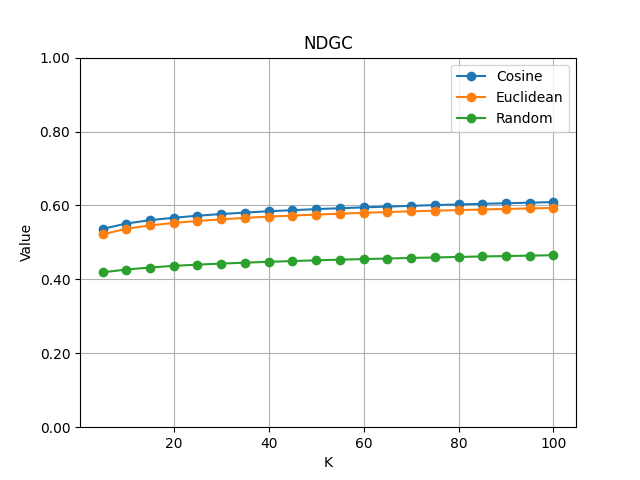}
    \caption{NDGC@K with different retrieval methods.}
    \label{fig:ndgc}
\end{figure}

\begin{figure}[hbt!]
    \centering
    \includegraphics[width=\linewidth]{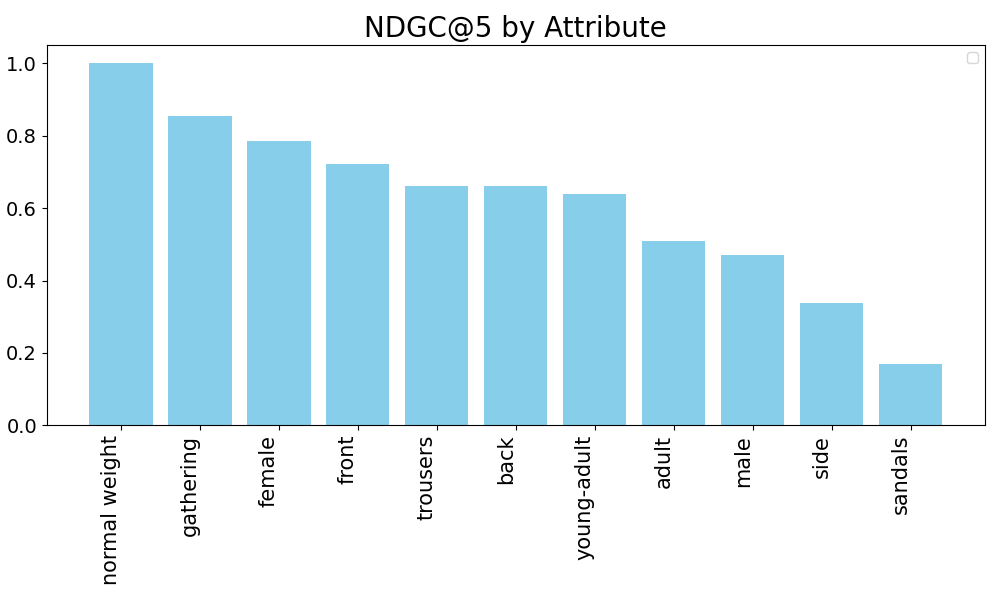}
    \caption{Top NDGC@5 results by attribute.}
    \label{fig:ndgc_feat}
\end{figure}

The queries were built by sampling one description for each of the unique feature combinations, resulting in 2721 different queries.  We compare retrieval based on euclidean distance and on cosine similarity with a random baseline. The results are shown in Figure \ref{fig:ndgc}. Our best performing approach, obtains an average of 60\% NDGC, considerably higher than the random baseline. We also compute the NDCG@5 score per each individual attribute. The attributes that exhibit best performance are showcased in Figure \ref{fig:ndgc_feat}.

\section{CONCLUSIONS}
\label{sec:conclusions}
In this work, we showcased the potential applications of LLMs to generate rich, synthetic textual descriptions for motion sequences, to facilitate the alignment between text and motion. We used this approach in two settings: (i) to align textual representations of actions to the corresponding skeleton sequences on BABEL-60 action recognition benchmark \cite{babel} and (ii) to align outward appearance descriptions of persons to their walking patterns on DenseGait \cite{cosma22gaitformer}. Aligning appearance descriptions to gait is a new research direction in the field, and we presented proof-of-concept results that appearance can be indicative of a persons unconscious behaviour in the form of walking. Our method presents promising results in this area, motivating future research of using LLMs to provide general and diverse supervisory signal to understanding motion sequences.

\section*{ACKNOWLEDGEMENTS}
This work was partly supported by the NXP PhD Student Grants, by the Google IoT/Wearables Student Grants and by the Keysight Master Research Sponsorship.

{\small
\bibliographystyle{ieee}
\bibliography{refs}

\begin{thebibliography}{10}\itemsep=-1pt

\bibitem{teach}
N.~Athanasiou, M.~Petrovich, M.~J. Black, and G.~Varol.
\newblock Teach: Temporal action composition for 3d humans.
\newblock In {\em 2022 International Conference on 3D Vision (3DV)}, pages 414--423, 2022.

\bibitem{betker2023improving}
J.~Betker, G.~Goh, L.~Jing, T.~Brooks, J.~Wang, L.~Li, L.~Ouyang, J.~Zhuang, J.~Lee, Y.~Guo, et~al.
\newblock Improving image generation with better captions.
\newblock {\em Computer Science. https://cdn. openai. com/papers/dall-e-3. pdf}, 2:3, 2023.

\bibitem{bucur2023utilizing}
A.-M. Bucur.
\newblock Utilizing chatgpt generated data to retrieve depression symptoms from social media.
\newblock {\em arXiv preprint arXiv:2307.02313}, 2023.

\bibitem{catruna2023gaitpt}
A.~Catruna, A.~Cosma, and E.~Radoi.
\newblock Gaitpt: Skeletons are all you need for gait recognition.
\newblock {\em arXiv preprint arXiv:2308.10623}, 2023.

\bibitem{cosma22gaitformer}
A.~Cosma and E.~Radoi.
\newblock Learning gait representations with noisy multi-task learning.
\newblock {\em Sensors}, 22(18), 2022.

\bibitem{catruna2024paradox}
A.~Cătrună, A.~Cosma, and E.~Rădoi.
\newblock The paradox of motion: Evidence for spurious correlations in skeleton-based gait recognition models, 2024.

\bibitem{ESTEVAM2021159}
V.~Estevam, H.~Pedrini, and D.~Menotti.
\newblock Zero-shot action recognition in videos: A survey.
\newblock {\em Neurocomputing}, 439:159--175, 2021.

\bibitem{action2motion}
C.~Guo, X.~Zuo, S.~Wang, S.~Zou, Q.~Sun, A.~Deng, M.~Gong, and L.~Cheng.
\newblock Action2motion: Conditioned generation of 3d human motions.
\newblock In {\em Proceedings of the 28th ACM International Conference on Multimedia}, MM '20, page 2021–2029, New York, NY, USA, 2020. Association for Computing Machinery.

\bibitem{jia2021scaling}
C.~Jia, Y.~Yang, Y.~Xia, Y.-T. Chen, Z.~Parekh, H.~Pham, Q.~Le, Y.-H. Sung, Z.~Li, and T.~Duerig.
\newblock Scaling up visual and vision-language representation learning with noisy text supervision.
\newblock In {\em International conference on machine learning}, pages 4904--4916. PMLR, 2021.

\bibitem{jiang2023mistral}
A.~Q. Jiang, A.~Sablayrolles, A.~Mensch, C.~Bamford, D.~S. Chaplot, D.~de~las Casas, F.~Bressand, G.~Lengyel, G.~Lample, L.~Saulnier, L.~R. Lavaud, M.-A. Lachaux, P.~Stock, T.~L. Scao, T.~Lavril, T.~Wang, T.~Lacroix, and W.~E. Sayed.
\newblock Mistral 7b, 2023.

\bibitem{actiongpt}
S.~S. Kalakonda, S.~Maheshwari, and R.~K. Sarvadevabhatla.
\newblock Action-gpt: Leveraging large-scale language models for improved and generalized action generation.
\newblock In {\em 2023 IEEE International Conference on Multimedia and Expo (ICME)}, pages 31--36, 2023.

\bibitem{DBLP:journals/corr/KayCSZHVVGBNSZ17}
W.~Kay, J.~Carreira, K.~Simonyan, B.~Zhang, C.~Hillier, S.~Vijayanarasimhan, F.~Viola, T.~Green, T.~Back, P.~Natsev, M.~Suleyman, and A.~Zisserman.
\newblock The kinetics human action video dataset.
\newblock {\em CoRR}, abs/1705.06950, 2017.

\bibitem{uae}
X.~Li and J.~Li.
\newblock Angle-optimized text embeddings, 2023.

\bibitem{li-etal-2023-synthetic}
Z.~Li, H.~Zhu, Z.~Lu, and M.~Yin.
\newblock Synthetic data generation with large language models for text classification: Potential and limitations.
\newblock In H.~Bouamor, J.~Pino, and K.~Bali, editors, {\em Proceedings of the 2023 Conference on Empirical Methods in Natural Language Processing}, pages 10443--10461, Singapore, Dec. 2023. Association for Computational Linguistics.

\bibitem{mteb}
N.~Muennighoff, N.~Tazi, L.~Magne, and N.~Reimers.
\newblock {MTEB}: Massive text embedding benchmark.
\newblock In A.~Vlachos and I.~Augenstein, editors, {\em Proceedings of the 17th Conference of the European Chapter of the Association for Computational Linguistics}, pages 2014--2037, Dubrovnik, Croatia, May 2023. Association for Computational Linguistics.

\bibitem{actor}
M.~Petrovich, M.~J. Black, and G.~Varol.
\newblock Action-conditioned 3d human motion synthesis with transformer vae.
\newblock In {\em 2021 IEEE/CVF International Conference on Computer Vision (ICCV)}, pages 10965--10975, 2021.

\bibitem{temos}
M.~Petrovich, M.~J. Black, and G.~Varol.
\newblock Temos: Generating diverse human motions from textual descriptions.
\newblock In S.~Avidan, G.~Brostow, M.~Ciss{\'e}, G.~M. Farinella, and T.~Hassner, editors, {\em Computer Vision -- ECCV 2022}, pages 480--497, Cham, 2022. Springer Nature Switzerland.

\bibitem{petrovich2024multi}
M.~Petrovich, O.~Litany, U.~Iqbal, M.~J. Black, G.~Varol, X.~B. Peng, and D.~Rempe.
\newblock Multi-track timeline control for text-driven 3d human motion generation.
\newblock {\em arXiv preprint arXiv:2401.08559}, 2024.

\bibitem{babel}
A.~R. Punnakkal, A.~Chandrasekaran, N.~Athanasiou, A.~Quiros-Ramirez, and M.~J. Black.
\newblock Babel: Bodies, action and behavior with english labels.
\newblock In {\em Proceedings of the IEEE/CVF Conference on Computer Vision and Pattern Recognition (CVPR)}, pages 722--731, June 2021.

\bibitem{qin2024diffusiongpt}
J.~Qin, J.~Wu, W.~Chen, Y.~Ren, H.~Li, H.~Wu, X.~Xiao, R.~Wang, and S.~Wen.
\newblock Diffusiongpt: Llm-driven text-to-image generation system, 2024.

\bibitem{clip}
A.~Radford, J.~W. Kim, C.~Hallacy, A.~Ramesh, G.~Goh, S.~Agarwal, G.~Sastry, A.~Askell, P.~Mishkin, J.~Clark, G.~Krueger, and I.~Sutskever.
\newblock Learning transferable visual models from natural language supervision.
\newblock In M.~Meila and T.~Zhang, editors, {\em Proceedings of the 38th International Conference on Machine Learning}, volume 139 of {\em Proceedings of Machine Learning Research}, pages 8748--8763. PMLR, 18--24 Jul 2021.

\bibitem{sahu-etal-2022-data}
G.~Sahu, P.~Rodriguez, I.~Laradji, P.~Atighehchian, D.~Vazquez, and D.~Bahdanau.
\newblock Data augmentation for intent classification with off-the-shelf large language models.
\newblock In {\em Proceedings of the 4th Workshop on NLP for Conversational AI}, pages 47--57, Dublin, Ireland, May 2022. Association for Computational Linguistics.

\bibitem{unipt}
Z.~Shao, X.~Zhang, C.~Ding, J.~Wang, and J.~Wang.
\newblock Unified pre-training with pseudo texts for text-to-image person re-identification.
\newblock In {\em 2023 IEEE/CVF International Conference on Computer Vision (ICCV)}, pages 11140--11150, Los Alamitos, CA, USA, oct 2023. IEEE Computer Society.

\bibitem{2sAGCN}
L.~Shi, Y.~Zhang, J.~Cheng, and H.~Lu.
\newblock Two-stream adaptive graph convolutional networks for skeleton-based action recognition.
\newblock In {\em Proceedings of the IEEE/CVF Conference on Computer Vision and Pattern Recognition (CVPR)}, June 2019.

\bibitem{cvae}
K.~Sohn, H.~Lee, and X.~Yan.
\newblock Learning structured output representation using deep conditional generative models.
\newblock In C.~Cortes, N.~Lawrence, D.~Lee, M.~Sugiyama, and R.~Garnett, editors, {\em Advances in Neural Information Processing Systems}, volume~28. Curran Associates, Inc., 2015.

\bibitem{motionclip}
G.~Tevet, B.~Gordon, A.~Hertz, A.~H. Bermano, and D.~Cohen-Or.
\newblock Motionclip: Exposing human motion generation to clip space.
\newblock In {\em Computer Vision – ECCV 2022: 17th European Conference, Tel Aviv, Israel, October 23–27, 2022, Proceedings, Part XXII}, page 358–374, Berlin, Heidelberg, 2022. Springer-Verlag.

\bibitem{veselovsky2023generating}
V.~Veselovsky, M.~H. Ribeiro, A.~Arora, M.~Josifoski, A.~Anderson, and R.~West.
\newblock Generating faithful synthetic data with large language models: A case study in computational social science, 2023.

\bibitem{wang2023visionllm}
W.~Wang, Z.~Chen, X.~Chen, J.~Wu, X.~Zhu, G.~Zeng, P.~Luo, T.~Lu, J.~Zhou, Y.~Qiao, et~al.
\newblock Visionllm: Large language model is also an open-ended decoder for vision-centric tasks.
\newblock {\em arXiv preprint arXiv:2305.11175}, 2023.

\bibitem{Yuan2024-al}
J.~Yuan, R.~Tang, X.~Jiang, and X.~Hu.
\newblock Large language models for healthcare data augmentation: An example on {Patient-Trial} matching.
\newblock {\em AMIA Annu Symp Proc}, 2023:1324--1333, Jan. 2024.

\bibitem{zhao2023survey}
W.~X. Zhao, K.~Zhou, J.~Li, T.~Tang, X.~Wang, Y.~Hou, Y.~Min, B.~Zhang, J.~Zhang, Z.~Dong, Y.~Du, C.~Yang, Y.~Chen, Z.~Chen, J.~Jiang, R.~Ren, Y.~Li, X.~Tang, Z.~Liu, P.~Liu, J.-Y. Nie, and J.-R. Wen.
\newblock A survey of large language models, 2023.

\bibitem{rot6d}
Y.~Zhou, C.~Barnes, J.~Lu, J.~Yang, and H.~Li.
\newblock On the continuity of rotation representations in neural networks.
\newblock In {\em 2019 IEEE/CVF Conference on Computer Vision and Pattern Recognition (CVPR)}, pages 5738--5746, 2019.

\bibitem{LLM4IRSurvey}
Y.~Zhu, H.~Yuan, S.~Wang, J.~Liu, W.~Liu, C.~Deng, H.~Chen, Z.~Dou, and J.-R. Wen.
\newblock Large language models for information retrieval: A survey.
\newblock {\em CoRR}, abs/2308.07107, 2023.

\end{thebibliography}
}

\end{document}